\documentclass{IEEEcsmag}

\PassOptionsToPackage{hyphens}{url}\usepackage{hyperref}
\expandafter\def\expandafter\UrlBreaks\expandafter{\UrlBreaks\do\/\do\*\do\-\do\~\do\''\do\"\do\-}

\usepackage{mathabx}
\usepackage{soulpos}
\usepackage{xcolor}
\usepackage{flushend}
\usepackage[backend=bibtex, style=ieee, sorting=nyt]{biblatex}
\usepackage{comment}
\usepackage{amsmath}
\usepackage{mdframed}
\usepackage{ragged2e}
\usepackage{multicol}

\begin{document}


\title{Turing's Test, a Beautiful Thought Experiment}

\author{Bernardo Gonçalves$\,$\textsuperscript{\href{https://orcid.org/0000-0003-2794-8478}{\includegraphics[scale=0.08]{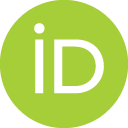}}}$\,$}
\affil{University of São Paulo, São Paulo, Brazil, \& King's College, Cambridge, UK}

\vspace{5pt}
\markboth{}{}

\begin{abstract}\looseness-1
In the wake of the latest trends of artificial intelligence (AI), there has been a resurgence of claims and questions about the Turing test and its value, which are reminiscent of decades of practical ``Turing'' tests. If AI were quantum physics, by now several ``Schr\"odinger's'' cats would have been killed. 
It is time for a historical reconstruction of Turing's beautiful thought experiment. This paper presents a wealth of evidence, including new archival sources, and gives original answers to several open questions about Turing's 1950 paper, including its relation with early AI. 
\end{abstract}

\maketitle

\noindent
Since the early 1990s, Turing's test has been used for publicity purposes as a practical experiment, and has been aptly criticized \cite{shieber1994,shieber1994b,vardi2014}. It has been the whipping boy of AI \cite{hayes1995}, cognitive sciences and analytic philosophy (cf.~\cite{shieber2004}), and increasingly, with the rise of AI, the humanities and social sciences \cite{brynjolfsson2022}.  
It is not uncommon for critics from all these fields to take Turing's test literally, while ignoring parts of his 1950 text. It is often assumed that he was promoting deception as a criterion for intelligence and/or proposing a crucial experiment to establish the existence of machine intelligence.  
Now, with the latest trends of AI technology, publications ask whether the Turing test can be a ``benchmark'' for AI \cite{biever2023}, and whether it is ``dead'' \cite{wells2023}. 
Based on recent primary research \cite{goncalves2023experiment,goncalves2023galileo,goncalves2023irony,goncalves2023lovelace,goncalves2023argument}, this paper presents a new perspective on Turing's test.  

\begin{figure}
\includegraphics[width=0.47\textwidth]{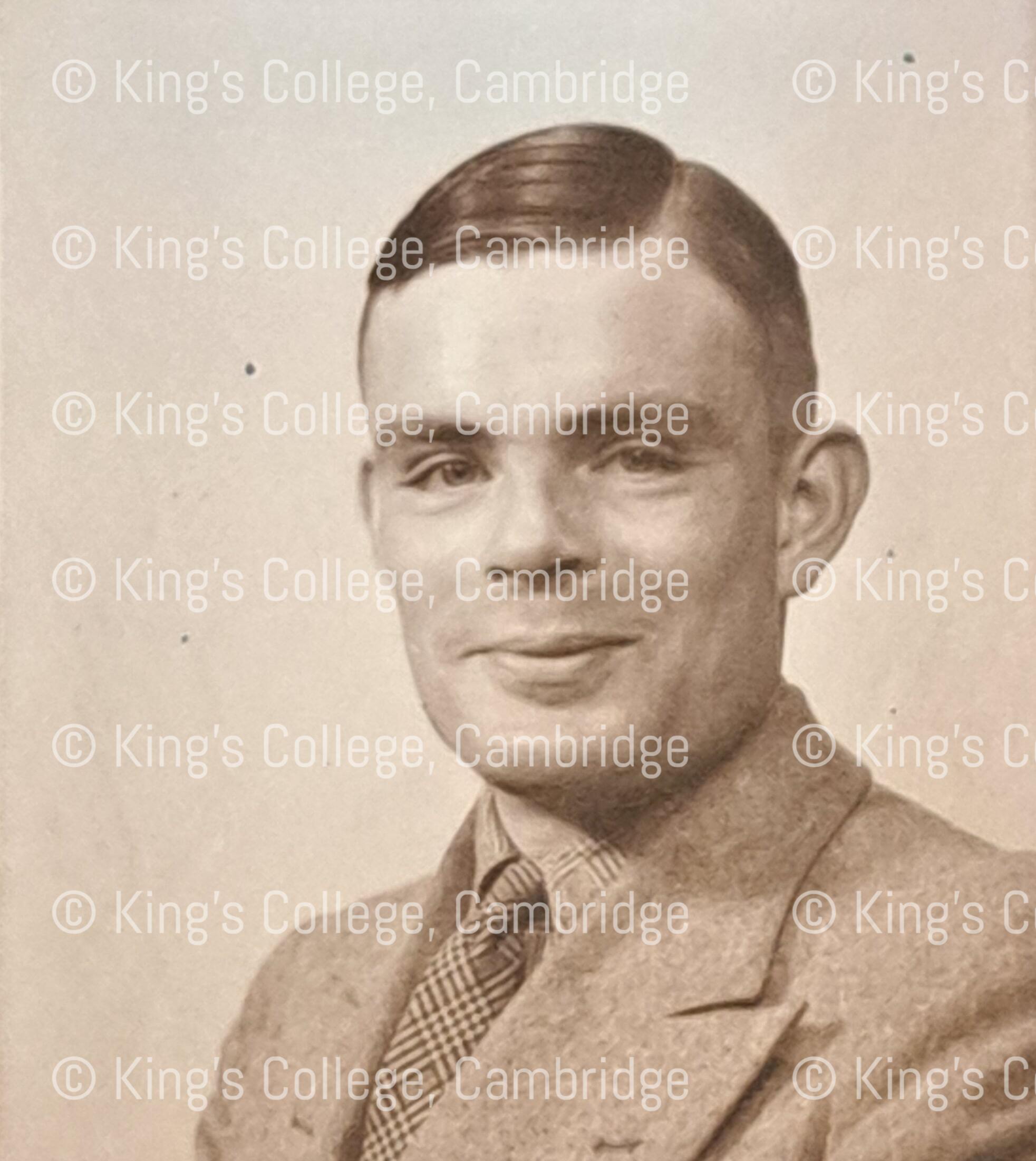}
\caption{Alan Turing (1912-1954). Photographs of Alan Turing, copyright The Provost and Scholars of King's College Cambridge 2023. Archives Centre, King's College, Cambridge, AMT/K/7/12. Reproduced with permission.}
\label{fig:turing}
\vspace{-15pt}
\end{figure}

A structural reading of Turing's 1950 paper is presented. Newly discovered archival sources are introduced and Turing's concept of imitation is examined. It is argued that his discursive presentation of the imitation game in various versions, as opposed to a well-defined, controlled experiment, conforms to what the physicist and historian Ernst Mach called ``the basic method of thought experiments'' \cite{mach1897}. The historical conditions of Turing's proposal are reconstructed from a controversy about mind and computing machine in England, contextualizing his requirement of gender imitation and conversational performance as success criteria for machine intelligence. Turing's responses to critics and propaganda for machine intelligence are highlighted. Finally, the question of the value of Turing's test and its relation to the history of AI is revisited.

\vspace{-6pt}
\section{WHAT IS THE TURING TEST?}\label{sec:test}
\noindent
In 1950, Alan Turing (Fig.~\ref{fig:turing})
published ``Computing Machinery and Intelligence'' \cite{turing1950}, the second of his three seminal papers.%
\footnote{The other two being ``On Computable Numbers'' \cite{turing1936} and ``The Chemical Basis of Morphogenesis'' \cite{turing1952b}.} 
The text has 28 pages, divided in seven sections, \S1-\S7. Three main logical steps can be identified in his argument: the \emph{proposal} (\S1-\S3, in $3+$ pp.), the \emph{science} (\S4-\S5, in 6 pp.), and the \emph{discussion} (\S6-\S7, in 18+ pp.). This structure, with its thematic order and size distribution, can be revealing about the nature of Turing's paper and argument.

The \emph{proposal} sought to replace with the imitation game the question ``Can machines think?,'' which he considered ``too meaningless to deserve discussion'' (p.~442; Turing was coming from unstructured multidisciplinary debates in two editions of a seminar, ``Mind and Machine,'' held at the Philosophy Department of Manchester University in October and December, 1949.)%
\footnote{Of the December edition, a participant wrote: ``I wish you had been with us a few days ago we had an amusing evening discussion with Thuring [\emph{sic}], Williams, Max Newman, Polyani [\emph{sic}], Jefferson, J Z Young \& myself $\hdots$ An electronic analyser and a digital computer (universal type) might have sorted the arguments out a bit.'' Christmas postcard from Jules Y. Bogue to Warren McCulloch, c. December, 1949. American Philosophical Society, W. S. McCulloch Papers, Mss.B.M139\_005. Thanks to Jonathan Swinton for this archival finding.} 
The purpose of the proposal was to change the common meaning of the word ``machine'' (e.g., a steam engine, a bulldozer) in light of the new mathematical \emph{science} of ``universal'' digital computing. The imitation game would allow for a grounded \emph{discussion} of ``machine'' and ``thinking,'' seeking to expand the meaning of ``thinking'' and detach it from the human species, much as the meaning of ``universe'' was once detached from the Earth, but also as a critique of anthropocentrism.

In 1950, one of the OED definitions of ``machine'' was:%
\footnote{New English Dictionary. Oxford, Vol. VI, Part II, M-N, p.~7.} 
``a combination of parts moving mechanically as contrasted with a being having life, consciousness and will $\hdots$ Hence applied to a person who acts merely from habit or obedience to a rule, without intelligence, or to one whose actions have the undeviating precision and uniformity of a machine.'' 
Thus, by definition, common sense did not allow the meanings of ``machine'' and ``thinking'' to overlap. Despite Turing's emphasis in his opening paragraph that he did not intend to discuss how these words were ``commonly used'' (p.~433), the hostility to his proposal can be seen from one of the first reactions, from a participant in the 1949 Manchester seminars, who quoted the above OED definition to appeal to common sense \cite[p.~149]{mays1952}.

The new question, which Turing considered to have a ``more accurate form'' \cite[p.~442]{turing1950}, would be based on a vivid image, his ``criterion for `thinking'$\,$'' (p.~436), which he called interchangeably the ``imitation game'' and his ``test.''%
\footnote{Turing referred to his ``test'' four times --- in pp. 446--447, 454. He also referred to it as an ``experiment'' --- once on p. 436, twice on p. 455, and twice again on p. 457.} 
The new question was whether a machine playing A, the deceiver, could imitate a woman, a man, a human being, or a different machine playing B, the assistant, in a remotely played conversation game to pass as B in the eyes of an average human interrogator playing C, the judge.

\sethlcolor{yellow}
However, the details and exact conditions of the imitation game as an experiment slipped through Turing's text in a series of variations that defies interpretation. A close reading of the text identifies four different conditions of the game with respect to players A-B, namely, man-woman (p.~433), machine-woman (p.~434), machine-machine (pp.~441, 451-452), and machine-man (p.~442). 
These different conditions relate to four variants of the ``new'' question that Turing posed to replace his ``original'' question (see Box~1). 
In addition to varying the species (types) of the players, he also increased the storage and speed of the machine and provided it with a hypothetically appropriate program ($Q^{\prime\prime\prime}$), and suggested a base time for the interrogation session ($Q^{\prime\prime\prime\prime}$). Other seemingly relevant parameters were not mentioned, such as the number of interrogators used to arrive at a statistically sound conclusion, although their profile is mentioned --- they should be ``average'' ---, and later reiterated --- they ``should not be expert about machines.''%
\footnote{``Can automatic calculating machines be said to think?,'' Broadcast on BBC Third Programme, 14 and 23 Jan. 1952. Archives Centre, King's College, Cambridge, AMT/B/6.}
\sethlcolor{pink}

\begin{figure*}[t]
\begin{mdframed}[backgroundcolor=blue!9] 
\textbf{BOX 1}\vspace{-5pt}\\
\noindent\rule{\columnwidth}{0.3pt}\vspace{1pt}\\
\begin{Large}
\textbf{The various questions and conditions of Turing's test}%
\vspace{-6pt}\\
\end{Large}
\noindent\rule{\columnwidth}{0.3pt}\vspace{-28pt}\\
\begin{multicols}{2}
\noindent
$Q$: ``I propose to consider the question, `Can machines think?'$\,$'' (p.~433)
\vspace{3pt}

\noindent
$Q^{\prime}$: ``We now ask the question, `What will happen when a machine takes the part of A in this game?’ Will the interrogator decide wrongly as often when the game is played like this [machine-woman] as he does when the game is played between a man and a woman? These questions replace our original, `Can machines think?'$\,$'' (pp.~433-434)
\vspace{3pt}

\noindent
$Q^{\prime\prime}$: ``There are already a number of digital computers in working order, and it may be asked, ‘Why not try the experiment straight away? It would be easy to satisfy the conditions of the game. A number of interrogators could be used, and statistics compiled to show how often the right identification was given.’ The short answer is that we are not asking whether all digital computers would do well in the game nor whether the computers at present available would do well, but whether there are imaginable computers which would do well.'' (p.~436)
\vspace{3pt}

\noindent
$Q^{\prime\prime\prime}$: ``It was suggested tentatively that the question [$Q$] 
should be replaced by [$Q^{\prime\prime}$] $\hdots$ 
But in view of the universality property we see that either of these questions is equivalent to this,  
`Let us fix our attention on one particular digital computer C. Is it true that by modifying this computer to have an adequate storage, suitably increasing its speed of action, and providing it with an appropriate programme, C can be made to play satisfactorily the part of A in the imitation game, the part of B being taken by a man?'$\,$'' (p.~442)
\vspace{3pt}

\noindent
$Q^{\prime\prime\prime\prime}$: ``I believe that in about fifty years' time it will be possible to programme computers, with a storage capacity of about $10^9$, to make them play the imitation game so well that an average interrogator will not have more than 70 per cent. chance of making the right identification after five minutes of questioning.'' (p.~442)
\end{multicols}
\vspace{-6pt}
\end{mdframed}
\vspace{-12pt}
\end{figure*}

\sethlcolor{pink}
Version $Q^{\prime\prime\prime\prime}$ of the test appears at the beginning of \S6 of the 1950 paper. As we will see shortly, this is the version that Turing most directly associates with the idea of a future experiment. As the least underspecified version of the test (cf.~Box~1), it has been the one picked up by promoters of practical ``Turing'' tests. In that passage, Turing expresses his belief that in ``about fifty years'' an ``average interrogator'' would miss the identification in at least 30\% of the test sessions. 

Two sentences later, Turing states a second belief: 

\begin{quotation}\noindent
I believe that at the end of the century the use of words and general educated opinion will have altered so much that one will be able to speak of machines thinking without expecting to be contradicted. 
\cite[p.~442]{turing1950}
\end{quotation}

\noindent
Once again we are brought to a crucial moment at the end of the century. But this second stated belief neatly reformulates the first, to which it is almost juxtaposed, and seems to reveal in common language what is meant by the rhetoric of achieving 30\% of misidentification in the imitation game: it expresses in rough round numbers the experience of living in the culture he envisions, where ``one will be able to speak of machines thinking without expecting to be contradicted.''  
Could such a cultural shift come with the future of digital computing? How remote was it? These discursive questions are arguably the real questions he addresses. What about running his test? 

Having gone through nine objections to machine intelligence discussed on the basis of the imitation game, we come to the crux: Turing's reference to ``experiment'' at the beginning of his \S7, ``Learning Machines.'' 
This passage, which comes after his revisiting of Lady Lovelace's Objection, has received little attention:

\begin{quotation}\noindent
These last two paragraphs do not claim to be convincing arguments. They should rather be described as `recitations tending to produce belief.' 
\end{quotation}

\noindent
[Continues]

\begin{quotation}\noindent
The only really satisfactory support that can be given for the view expressed at the beginning of \S6 [his two stated beliefs] 
will be that provided by waiting for the end of the century and then doing the experiment described.  
\end{quotation}

\noindent
[Concludes]

\begin{quotation}\noindent
But what can we say in the meantime? What steps should be taken now if the experiment is to be successful? 
\cite[p.~455]{turing1950}
\end{quotation}

\noindent
This passage, in three sentences followed by two rhetorical questions, sums up the sophistication of Turing's rhetoric. The suggestion to run the experiment comes juxtaposed with his indication that he is actively engaged in propaganda (``recitations tending to produce belief''). Again, he pushes ``the experiment'' to ``the end of the century,'' but now lands in the present, in a call to arms for research into ``learning machines'' so that the experiment --- an iconic representation of the change he expects to see in talk of ``machines thinking'' --- could be successful. 

The contrast between what he proposes for the future and for the present is revealing. If the research is done as he suggests, by the time ``the experiment'' is to be conducted, he expects the machines to be so advanced that talk of ``machines thinking'' will be commonplace. In his major 1948 ``Intelligent Machinery'' report,%
\footnote{Archives Centre, King's College, Cambridge, AMT/C/11.}
 the rhetoric of a crucial experiment does not appear. It was rather ``the actual production of the machines'' that ``would probably have some effect'' in convincing critics and opponents, because ``the idea of `intelligence' is itself emotional rather than mathematical'' (p.~3). 
But in the crucial year of 1949, as we will see later, Turing faced the strongest wave of opposition from contemporaries, leading to his test.

\begin{figure*}[t]
\begin{mdframed}[backgroundcolor=yellow!20] 
\textbf{BOX 2}\vspace{-5pt}\\
\noindent\rule{\columnwidth}{0.3pt}\vspace{1pt}\\
\begin{Large}
\textbf{Turing's mathematical concept of imitation}\vspace{-6pt}\\
\end{Large}
\noindent\rule{\columnwidth}{0.3pt}\vspace{-27pt}\\
\begin{multicols}{2}
\noindent
Dear Miss Worsley,\\
\indent
I was interested in your work on the relation between computers and
Turing machines. I think it would be better though if you could try and find a realtion [\emph{sic}] between T machines and infinite computers, rahter [\emph{sic}] than between finite T machines and computers. The relation that you suggest is rather too trivial. The fact is that the motions of either a finite T machine or a finite computer are ultimately periodic, and therefore any sequence computed by them is ultimately periodic. It is easy therefore in theory to make one \emph{imitate} the other, though the size of the \emph{imitating} machine will (if this technique is adopted) have to be of the order of the exponential of the size of the \emph{imitated} machine. Probably your methods could prove that this exponential relation could be reduced to a multiplicative factor. \\
\indent
Yours sincerely, A. M. Turing%
\footnote{
Turing to B. H. Worsley, June 11, 1951, typeset; emphasis added. Unpublished writings of Alan Turing, copyright The Provost and Scholars of King’s College Cambridge 2023. B.H. Worsley Collection, Archives Center, National Museum of American History, Smithsonian Institution. Quoted with permission. Thanks to Mark Priestley for this archival finding.}
\end{multicols}
\end{mdframed}
\vspace{-15pt}
\end{figure*}

\vspace{-10pt}
\section{IMITATION: FROM 1936 TO 1950}\label{sec:imitation}

\sethlcolor{yellow}
Because the machine must imitate stereotypes of what it is not, Turing's proposal has often been criticized for encouraging fakes and tricks. But this view is related to the reading that Turing would have meant his test as a practical experiment. 
Such a literal reading of Turing's test misses the point of his use of irony \cite{goncalves2023irony}, and misses the fact that his notion of imitation in 1950 was largely in continuity with his 1936 paper \cite{turing1936}. This was hinted at in the words of the director of the National Physical Laboratory in a BBC broadcast in late 1946;%
\footnote{``$\hdots$about twelve years ago, a young Cambridge Mathematician by the name of Turing, wrote a paper which appeared in one of the mathematical journals, in which he worked out by strict logical principles, how far a machine could be imagined which would \emph{imitate} the processes of thought'' (emphasis added). \emph{The Listener}, Nov. 14, 1946, p.~663.}
and both in Turing's lecture in early 1947 and in his NPL report in mid-1948;%
\footnote{``Lecture to L.M.S. Feb. 20 1947'' and ``Intelligent Machinery.'' Archives Centre, King's College, Cambridge, AMT/B/1 and AMT/C/11.}
and as newly discovered correspondence with the Mexican-Canadian computer pioneer Beatrice Worsley (1921-1972) helps to clarify (see Box~2). 

In his letter to Worsley, Turing seems to be more interested in the relations between ``the motions'' of Turing machines and infinite computers, whose behavior can be non-periodic. 
Perhaps he thought of the living human brain as an infinite computer, in the sense that it has a continuous interface with its environment, which constantly intervenes and changes its logical structure.%
\footnote{``Intelligent Machinery'' (\emph{op. cit.}).}
Now, the imitation game puts into empirical form the relation between digital computers, whose behavior is ultimately periodic, and the behavior of the human players. Can the behavior of their brains be approximated by a digital computer? Turing pursued this question. For his May 1951 broadcast, he wrote: 
``the view which I hold myself, that it is not altogether unreasonable to describe digital computers as brains $\hdots$ 
If it is accepted that real brains, as found in animals, and in particular in men, are a sort of machine it will follow that our digital computer, suitably programmed, will behave like a brain.''%
\footnote{``Can digital computers think?,'' broadcast on BBC Third Programme, 15 May 1951. Archives Centre, King's College, Cambridge, AMT/B/5.}

\begin{figure*}[t]
\begin{mdframed}[backgroundcolor=pink!45] 
\textbf{BOX 3}\vspace{-5pt}\\
\noindent\rule{\columnwidth}{0.3pt}\vspace{1pt}\\
\begin{Large}
\textbf{$\!$``$\!$...any \emph{sharp} line between what machine and brain can do will fail''$\!$}\vspace{-6pt}\\
\end{Large}
\noindent\rule{\columnwidth}{0.3pt}\vspace{-26pt}\\
\begin{multicols}{2}
Dear Miss Worsley, $\,\hdots$\\
\indent 
I do not think you will be able to find any clue to essential differences between brains and computing machines (if there are any), in neuron behaviour. So long as what we know about a neuron can be embodied in the description of stochastic processes, the behaviour of any mechanism embodying such neurons can, in principle, be calculated by a suitable enlarged and speeded up \st{Ferranti} [Mark II] machine.%
\footnote{``Ferranti'' is typed and erased, and `Mark II' (a version of the Manchester electronic computer) is added in pencil.} 
More accurately I should say that one can calculate random samples of its behaviour. I think any attempt to draw any \emph{sharp} line between what machine and brain can do will fail. I think it is largely a quantitative matter. Probably one needs immensely more storage capacity then [\emph{sic}] we have got, and possibly more than we shall ever have. Perhaps we may have enough capacity, but just won’t find an appropriate programme. Naturally one won’t make a man that way ever. It’ll just be another species of the thinking genus.\\
\indent Yours sincerely, A. M. Turing%
\footnote{
Turing to B. H. Worsley, circa June, 1951, Turing's emphasis. Credit for this source is exactly the same as for that of Box~2. Quoted with permission.}
\end{multicols}
\end{mdframed}
\vspace{-10pt}
\end{figure*}

\begin{figure}[t]
\includegraphics[width=.48\textwidth]{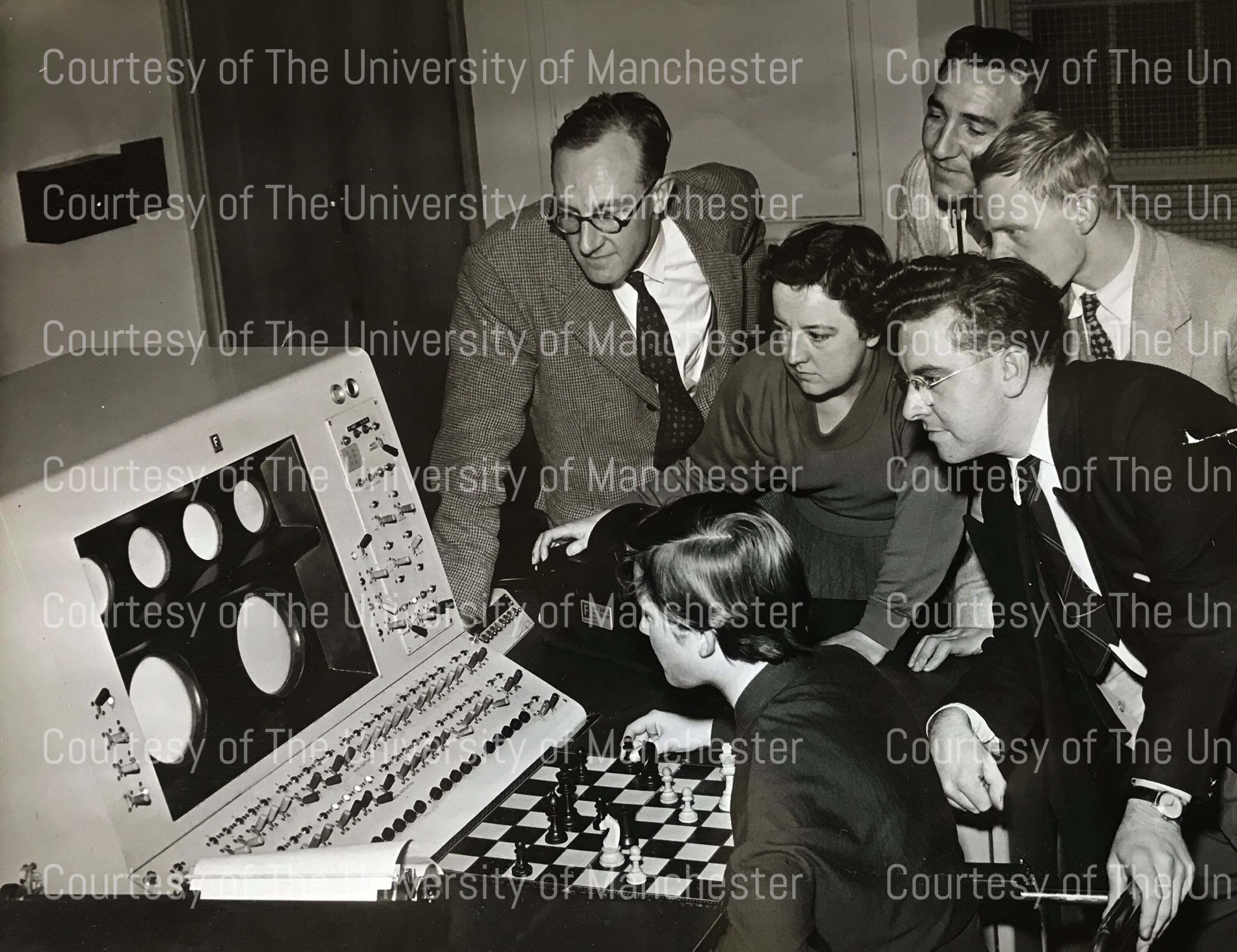}
\vspace{-10pt}
\caption{Console of Ferranti Mark I and a group with Turing's secretary at the Computing Machine Laboratory, Sylvia Robinson (née Wagstaff), pretending to play chess with the machine, c. 1955. Courtesy of The University of Manchester.} 
\label{fig:mark1}
\vspace{-15pt}
\end{figure}

Even if the human brain can only be compared to an infinite computer, could it not be simulated by a digital computer equipped with a sufficiently large memory? An excerpt of another newly discovered Turing letter to Worsley from mid-1951 can give more contour and provide further insight into Turing's views (see Box~3). A highlight in this excerpt is Turing's view that to the extent that the behavior of a neuron can be described as a stochastic process, it would be possible to ``calculate random samples'' of the mechanism that embodies the brain and then imitate it. 
An effective imitation of the brain by a machine would require knowledge of the anatomy and physiology of the brain to inspire an appropriate program, as well as much more storage and speed than was available to the Ferranti Mark I at the time (see Fig.~\ref{fig:mark1}). 
Another important element in the excerpt is Turing's point that, even if a thinking machine is possible, the relation he has in mind is not one of identity but one of analogy: ``It'll just be another species of the thinking genus.''

An original answer to the question of why design a test based on imitation, which can be seen as encouraging deception, is that imitation was actually Turing's fundamental principle of the new science of universal digital computing. He conceived his 1950 paper partly in continuity with his 1936 paper. Both were based on his core concepts of machine and imitation, i.e., what it takes for a machine to imitate another machine. A key difference that breaks the continuity is that, by 1950, he had generalized the logical architecture of his universal machine. It would not only follow instructions, but would also be able to acquire new cognitive skills by learning. Using Turing's 1948 language,%
\footnote{``Intelligent Machinery'' (\emph{op. cit.}).}
universality can be achieved by starting with an ``organized'' machine (1936),$\!$ or with an ``unorganized'' machine (1948/1950). Whereas in 1936 the machine would be given an a priori, well-defined and fixed table of instructions for each task, in 1950 it would also be able to perform a new task by changing its logical structure as a result of learning by experience, much as the brain does ``by changing its neuron circuits by the growth of axons and dendrites.''%
\footnote{Turing to Ross Ashby, circa November 19, 1946. British Library, Collection ``W. Ross Ashby: correspondence of W. Ross Ashby,'' Add MS 89153/26.}

The use of psychological tricks in practical ``Turing'' tests has little to do with Turing's 1950 proposal. In 1951, he warned: ``It would be quite easy to arrange the [machine's] experiences in such a way that they automatically caused the structure of the machine to build up into a previously intended form, and this would obviously be a gross form of cheating, almost on a par with having a man inside the machine.''%
\footnote{``Intelligent machinery, a heretical theory,'' a lecture given to ``51 Society'' at Manchester, c. 1951. Archives Centre, King's College, Cambridge, AMT/B/4.}
The ``human fallibility'' that Turing encouraged the machine to show was meant as a by-product of learning by experience \cite[p.~459]{turing1950}: 
``Another important result of preparing our machine for its part in the imitation game by a process of teaching and learning is that `human fallibility' is likely to be omitted [from the teaching] in a rather natural way, i.e., [learned] without special `coaching'.'' 
That is, for a machine to be a valid player of Turing's test, it cannot be specially prepared for it. This means that we have never seen a practical \emph{Turing} test.

\vspace{-10pt}
\section{THE METHOD OF THOUGHT EXPERIMENTS}\label{sec:mach}

The various rhetorical questions Turing posed, $Q^{\prime} \hdots Q^{\prime\prime\prime\prime}$, to replace the original question, $Q$, can be generalized as follows \cite{goncalves2023experiment}: 

Question $Q^\star$: could player A imitate intellectual stereotypes associated with player B's type successfully (well enough to deceive player C), despite A and B's physical differences? 

It has been largely unnoticed that the various questions instantiating $Q^\star$ can be viewed as following a case-control methodology, applied in two stages. 

\sethlcolor{yellow}
At the more obvious intra-game level, A plays the case, and B plays the control. However, at the inter-game level, two variants set the case (machine-woman and the machine-man) and the other two set the control (man-woman and the machine-machine). While the first two are open, creating suspense around the test, the latter two are resolved as follows. 
It is known that a man (A) can imitate gender stereotypes associated with a woman (B) to deceive an interrogator (C) despite their physical differences, as assumed in parlor games and as Turing illustrates on p. 434: ``My hair is shingled...''  
Further, regarding the machine-machine variant, it is also known that a digital computer (A), because of its universality property \cite[\S\S4, 5]{turing1950}, can imitate any discrete-state machine (B) and even a continuous one (p.~451), despite their physical differences. 

We can now explore how Turing's presentation of his test fits Mach's description of ``the basic method of thought experiments,'' which is variation, continuously if possible. 
Mach is the author of perhaps the most classic text on thought experiments in the modern scientific tradition \cite{mach1897}, in which he developed observations and insights based on examples from the history of modern physics, mathematics, and common sense experience. He wrote: ``By varying the conditions (continuously if possible), the scope of ideas (expectations) tied to them is extended: by modifying and specializing the conditions we modify and specialize the ideas, making them more determinate, and the two processes alternate'' (p.~139).
Mach illustrated his point with the process of discovery of universal gravitation: 

\begin{quotation}\noindent
A stone falls to the ground. Increase the stone’s distance from the earth, and it would go against the grain to expect that this continuous increase would lead to some discontinuity. Even at lunar distance the stone will not suddenly lose its tendency to fall. Moreover, big stones fall like small ones: the moon tends to fall to the earth. Our ideas would lose the requisite determination if one body were attracted to the other but not the reverse, thus the attraction is mutual and remains so with unequal bodies, for the cases merge into one another continuously $\hdots$ 
discontinuities are quite conceivable, but it is highly improbable that their existence would not have betrayed itself by some experience. \cite[pp. 138-139]{mach1897} 
\end{quotation}

\noindent
The conditions, i.e., the distance of the fall and the size of the stones, are continuously varied in the physicist's mind and eventually stretched to the celestial scale. Reciprocally, the concept of a celestial body, such as the Earth or the Moon, becomes interchangeable with the concept of a stone, and quite unequal stones can then become mutually attracted. 
The cases continuously merge into one another, and a conceptual integration is established that connects near-earth bodies to celestial bodies under a unified concept.

Turing's imitation game extended the scope of ideas and expectations established earlier in his 1936 paper, moving from machine-machine and restricted human-machine imitation in 1936%
\footnote{``We may compare a man in the process of computing a real number to a machine which is only capable of a finite number of conditions'' \cite[p.~231]{turing1936}.} 
to more general human-machine imitation in 1950.

To understand this better, let us take a brief look at Turing's 1948 report ``Intelligent Machinery'' (\emph{op. cit.}). 
In section (\S3) `Varieties of machinery,' he noted: ``All machinery can be regarded as continuous, but when it is possible to regard it as discrete it is usually best to do so.'' 
A brain, he wrote, ``is probably'' a `continuous controlling' machine, but in light of the digital nature of neural impulses, it ``is very similar to much discrete machinery.'' 
In section (\S6) ``Man as Machine,'' he referred to the imitation of ``any small part of a man'' by machines: ``A great positive reason for believing in the possibility of making thinking machinery is the fact that it is possible to make machinery to imitate any small part of a man'' (p.~420). 
In light of this, he argued: 
``One way of setting about our task of building a `thinking machine' would be to take a man as a whole and to try to replace all the parts of him by machinery.'' 
But Turing dismissed such a method as ``altogether too slow and impracticable,'' and later alluded to moral and aesthetic reasons as well.%
\footnote{For the May 1951 broadcast (\emph{op. cit.}), he wrote: ``I certainly hope and believe that no great efforts will be put into making machines with the most distinctively human, but non-intellectual characteristics such as the shape of the human body; it appears to me to be quite futile to make such attempts and their results would have something like the unpleasant quality of artificial flowers.'' }

We can now follow Turing's use of the method of continuous variation in the design of his imitation tests. 
The central question ($Q^\star$) Turing asks is whether the intellectual and cultural performances (the \emph{stereotypes})%
\footnote{Susan Sterrett first emphasized the importance of stereotypes in the imitation game \cite{sterrett2000}.} 
associated with woman, man, machine (the \emph{types}) could be imitated, and thus softly transposed. 
Note that for any arbitrarily chosen type, say, a woman, further specific subtypes can be continuously conceived and considered as varied conditions of the imitation game: women having a certain property, a subproperty, and so on. For any two arbitrarily chosen types, a new type can be conceived, whether as a specialization or a modification (``any small part of a man''). Because concepts are fluid abstractions, there is an evolving continuum of levels and types. 

The question across the various versions of the game can be posed this way: how does C's perception of A's performance against B's performance change as the game's conditions are (continuously) varied? 
Will it change if gendered verbal behavior is required as a subtype of human verbal behavior? 
Will it change if the machine's hardware is increased and/or its learning program is modified? 
For Turing, there is no conceptual discontinuity between the various conditions instantiating his thought experiment. It is often asked whether the imitation goal changes from the machine-woman test (p.~434) to the machine-man test (p.~442). Note that this open question vanishes under this interpretation of the test, which observes material and structural facts of Turing's text: he describes such goals only very loosely, and, more importantly, continuously varies the design of the game. His focus is on the power of universal digital computing to imitate and deconstruct arbitrary types, not on setting this or that type for the sake of promoting a particular test.  

From 1948 to 1952, Turing presented various imitation tests based on both the game of chess and conversation, even bringing back chess at the end of his 1950 paper (p.~460), and referring to his (``my'') various ``imitation tests'' in 1952.%
\footnote{`Can automatic calculating machines be said to think?', January 1952 (\emph{op. cit.}).} Why would he present various tests, as opposed to a well-defined, controlled experiment? This is a historically sound question, because it does not struggle with the materiality of Turing's texts and their chronological coherence. Nor does it erase some of his tests in favor of others, or overlook the historical conditions of his proposal. This paper has provided an answer by reconstructing what can be called Turing's use of the method of thought experiments, whose context will now be explored.

\vspace{-10pt}
\section{1949, THE CRUCIAL YEAR}\label{sec:1949}
\noindent
As is often the case with thought experiments, Turing proposed his test out of a controversy \cite{goncalves2023argument}. He was coming from his continuing disputes with the physicist and computer pioneer, Fellow of the Royal Society (FRS), Douglas Hartree (1897-1958), 
over the meaning of the newly existing digital computers, which had started in 1946 \cite{goncalves2023lovelace}. Now, in mid-1949, new opponents had arrived, most notably the neurosurgeon Geoffrey Jefferson (1886-1961), 
and the chemist and philosopher Michael Polanyi (1891-1976), 
both also FRS and based at the same institution as Turing, the University of Manchester, where Turing had spent a year as a Reader in the Department of Mathematics \cite{hodges1983}. 
These three thinkers challenged Turing's claims about the future possibilities and limitations of digital computers. 

In June 1949, Hartree published his \emph{Calculating Instruments and Machines} \cite{hartree1949}, in which Ada Lovelace's work was acknowledged seemingly for the first time by a twentieth-century computer pioneer \cite{goncalves2023lovelace}. Since November 1946, Hartree had been opposing the use of the term ``electronic brain.'' He wrote in a letter to the \emph{Times}: ``These machines can only do precisely what they are instructed to do by the operators who set them up.''%
\footnote{``The `Electronic Brain': A Misleading Term; No Substitute for Thought,'' \emph{Times}, November 7, 1946.}
Now in 1949, Hartree added strength to his argument by quoting the words of Ada Lovelace from the 1840s about Charles Babbage's machine: ``The Analytical Engine has no pretensions to originate anything $\hdots$ It can do \emph{whatever we know how to order it} to perform'' (her emphasis)  \cite[p.~70]{hartree1949}. 
Noting Hartree's anachronism in taking Lovelace's words out of their time and place, Turing further developed his earlier, 1947 response to Hartree's challenge,%
\footnote{`Lecture to L.M.S. Feb. 20 1947' (\emph{op. cit.}), p.~22.} 
now calling it ``(6) Lady Lovelace's objection'' \cite[p.~450]{turing1950}. Turing argued that intelligent behavior is the result of learning, a capability he had no problem attributing to future digital computers. 
He also questioned the implicit assumption of Hartree's challenge: ``Who can be certain that `original work' that he has done was not simply the growth of the seed planted in him by teaching, or the effect of following well-known general principles'' (p.~450). 
In the imitation game, Turing suggested, the interrogator would be able to evaluate the machine's ability to learn: ``The game (with the player B omitted) is frequently used in practice under the name of \emph{viva voce} to discover whether some one really understands something or has `learnt it parrot fashion'$\,$'' (p.~446). But then, what is player B doing in the imitation game? Following the 1949 events will suggest an answer.

On June 9, in London, Jefferson delivered his prestigious Lister Oration on ``The Mind of Mechanical Man,'' which was published in the debuting \emph{British Medical Journal} on June 25 \cite{jefferson1949}. His lecture was headlined in the \emph{Times} on June 10,%
\footnote{\mbox{``No Mind For Mechanical Man.'' \emph{Times},$\!$ 10 June 1949, p. 2.}}
emphasizing his claim that ``Not until a machine can write a sonnet or compose a concerto because of thoughts and emotions felt, and not by the chance fall of symbols, could we agree that machine equals brain'' (p.~1110).  
This rendered Turing's famous response: ``I do not think you can even draw the line about sonnets, though the comparison is perhaps a little bit unfair because a sonnet written by a machine will be better appreciated by another machine.''%
\footnote{``The Mechanical Brain.'' \emph{Times}, 11 June 1949, p. 4.} 
In October and December 1949, the two seminars on ``Mind and Machine'' were organized by Polanyi et al., and attended by Jefferson, Turing et al., at the Philosophy Department in Manchester \cite[][p.~275]{polanyi1958}. These seminar discussions, followed by Jefferson giving Turing an offprint of his Lister Oration,%
\footnote{This may have happened in the evening of the December meeting of the Manchester seminar (\emph{op. cit.}), when, according to a later letter from Jefferson to Ethel S. Turing, Turing and J.Z. Young went to dinner at Jefferson's house \cite[p.~xx]{sara1959}.}
which Turing read and marked with a pencil,%
\footnote{Off-print, ``The mind of mechanical man'' by Geoffrey Jefferson. Archives Centre, King's College, Cambridge, AMT/B/44.} 
led him to write his 1950 paper and propose his test \cite{goncalves2023argument}.

Jefferson had characterized intelligence as an emergent property of the animal nervous system. 
He emphasized that ``sex hormones introduce peculiarities of behaviour often as inexplicable as they are impressive'' (p.~1107). 
Because ``modern automata'' are not moved by male and female sex hormones, they could not exhibit such peculiarities to imitate the actions of animals or ``men.'' Specifically, he used a thought experiment to criticize Grey Walter's mechanical turtles by suggesting that gendered behavior is causally related to the physiology of sex hormones (\emph{ibid.}): 

\begin{quotation}
\noindent
[...It] should be possible to construct a simple animal such as a tortoise (as Grey Walter ingeniously proposed) that would show by its movements that it disliked bright lights, cold, and damp, and be apparently frightened by loud noises, moving towards or away from such stimuli as its receptors were capable of responding to. 
In a favourable situation the behaviour of such a toy could appear to be very lifelike --- so much so that a good demonstrator might cause the credulous to exclaim `This is indeed a tortoise.' I imagine, however, that another tortoise would quickly find it a puzzling companion and a disappointing mate. 
\end{quotation}

\noindent
In reaction to Grey Walter and his transgressive tortoises \cite{hayward2001},%
\footnote{Jefferson would attack Walter's automata again in speeches to learned societies in Manchester in 1953 \cite{jefferson1953} and 1956 \cite{jefferson1956} in the wake of Walter's \emph{The Living Brain} \cite{walter1953}.} 
Jefferson offered the image of a genuine individual of a species placed side by side with an artificial one to emphasize the latter's artificiality. The function of the genuine individual is to expose the artificiality of the impostor. The means of exposure is the failure to demonstrate the authentic (sexual) behavior of the original species. This can explain Turing's introduction of a (gendered) control player B, who appears in Turing's 1950 test, whose design was prompted by his reading of Jefferson, but not in Turing's 1948, 1951, and 1952 tests. 
In discussing ``(4) The Argument from Consciousness,'' Turing addressed Jefferson directly and quoted in full his conditions for agreeing ``that machine equals brain,'' including ``be warmed by flattery'' and ``be charmed by sex'' \cite[pp.~445-446]{turing1950}.%
\footnote{Jefferson's response was \cite[p.~73]{jefferson1953}: ``But there are those like Dr. Turing who believe that we have no right to deny self-consciousness to the machines since they fulfill the definition of mind as given above --- the ability to make choices.''}  
In discussing the ``(5) Argument from Various Disabilities'' (p.~447), Turing again addressed Jefferson (p.~450) and argued that to say that a machine could never ``fall in love'' or ``make someone fall in love with it'' was a flawed scientific induction from the capabilities of present machines.

Turing's test design may have been an ironic response to Jefferson's suggestion that gendered behavior is causally related to the physiology of male and female sex hormones. As a repressed homosexual \cite{hodges1983} and non-conformist in postwar England \cite{goncalves2023irony}, Turing might have been prepared to refer to a gender test. However, we have just seen that a basic version of this idea was actually available to him in the form of Jefferson's reaction to Walter's tortoises. 
Apart from Turing's 1950 paper, in which he is in direct dialogue with Jefferson, he links his views on machine intelligence to sex and gender in one other known source, again with Jefferson in the background. At the end of a letter to a friend written after the Wilmslow police challenged his testimony,%
\footnote{Turing to Norman Routledge, circa mid-Feb., 1952. Archives Centre, King's College, Cambridge, AMT/D/14a.}
he comments on the BBC broadcast of January 1952 (\emph{op. cit.}): ``Glad you enjoyed the broadcast. J. [Jefferson] certainly was rather disappointing though. I'm rather afraid that the following syllogism may be used by some in the future[:] Turing believes machines think. Turing lies with men. Therefore machines do not think.'' 
It remains to be explored Turing's choice of conversation for his test.

Surviving minutes of the ``Mind and Machine'' seminar held on October 27, 1949, were published in 2000 by a participant, Wolfe Mays \cite{mays2000}. 
In the first session, Polanyi presented a statement, ``Can the mind be represented by a machine?,''%
\footnote{Polanyi, Michael. Papers, Box 32, Folder 6, Hanna Holborn Gray Special Collections Research Center, University of Chicago Library.} 
which was a Gödelian argument that humans can do things that machines cannot. Although Turing had already addressed this argument in his 1947 lecture (\emph{op. cit.}), Polanyi's insistence may help explain Turing's inclusion of ``(3) The Mathematical Objection'' \cite[p.~444]{turing1950}.  Further, the minutes of the Manchester seminar show that Polanyi tried to distinguish the formal ``rules of the logical system'' from the informal ``rules which determine our own behaviour,'' and this helps explain Turing's inclusion of ``(8) The Argument from Informality of Behaviour'' (p.~452). Polanyi's argument came too late, as Turing had long been involved in such debates in the Moral Sciences Club at Cambridge University, both before and after World War II.%
\footnote{Minutes and other papers of the Moral Sciences Club, 1878--2018, Cambridge University Library, GBR/0265/UA/Min.IX.39-48$^\ast$, 56-6$^\ast$ etc.}

Years later \cite[p.~275]{polanyi1958}, Polanyi remembered ``a communication to a Symposium held on `Mind and Machine' at Manchester University in October, 1949,'' in which ``A.M. Turing has shown 
that it is possible to devise a machine which will both construct and assert as new axioms an indefinite sequence of Gödelian sentences.''%
\footnote{Polanyi added that ``this is foreshadowed'' in Turing's 1938 paper based on his Ph.D. thesis, ``Systems of Logic Based on Ordinals,'' \emph{J. London Math. Soc.} s2-45(1), 161-228.}  
Polanyi resumed, showing that he assimilated the punch: ``Any heuristic process of a routine character---for which in the deductive sciences the Gödelian process is an example---could likewise be carried out automatically.'' 
However, Polanyi used the same argument to dismiss the game of chess as a testbed for machine intelligence, noting: ``A routine game of chess can be played automatically by a machine, and indeed, all arts can be performed automatically to the extent to which the rules of the art can be specified.'' 

Chess, not conversation, had been Turing's chosen field to illustrate, develop, and test machine intelligence since at least February 1946.%
\footnote{``Proposed electronic calculator,'' February 1946. Archives Centre, King's College, Cambridge, AMT/C/32. On p.~16, Turing asks: ``Can the machine play chess?''}
In his 1948 `Intelligent Machinery' (\emph{op. cit.}, pp.~21-22), Turing had discussed a tradeoff between convenient and impressive intellectual fields for exploring machine intelligence. After discussing ``various games e.g. chess,'' he wrote: ``Of the above possible fields the learning of languages would be the most impressive, since it is the most human of these activities.'' 
However, he avoided language learning because it seemed ``to depend rather too much on sense organs and locomotion to be feasible,'' stuck with chess, and ended up describing a chess-based imitation game. 
Now, in October 1949, he saw chess being dismissed as unimpressive to make the case for machine intelligence because its rules could be specified. 

Some time later, probably around Christmas 1949, Turing read Jefferson's Lister Oration \cite{jefferson1949} and marked the passage quoting René Descartes (p.~1106), which starts: ``Descartes made the point, and a basic one it is, that a parrot repeated only what it had been taught and only a fragment of that; it never used words to express its own thoughts.'' Overall, Jefferson suggested `speech' to be the distinguishing feature of human intelligence compared to other kinds of animal intelligence: ``Granted that much that goes on in our heads is wordless $\hdots$ we certainly require words for conceptual thinking as well as for expression $\hdots$ It is here that there is the sudden and mysterious leap from the highest animal to man, and it is in the speech areas of the dominant hemisphere $\hdots$ that Descartes should have put the soul, the highest intellectual faculties'' (p.~1109).

Unlike chess, which is governed by definite rules, good performance in conversation cannot be easily specified. Turing's 1950 choice for ``the learning of languages'' as the intellectual field addressed in his test can be best understood as yet another concession to Jefferson and, in this case, to Polanyi as well.

Finally, Jefferson also argued that the nervous impulse is not a purely electrical phenomenon but also a chemical one that depends on the continuity of specific physical quantities (p.~1108). It would thus be incommensurable with the activity of a digital computer. In response, Turing formulated ``(7) The Argument from Continuity in the Nervous System'' \cite[p.~451]{turing1950}, in which he used the imitation game in its machine-machine version to neutralize that physical difference. A digital computer (a discrete-state machine) could imitate a differential analyzer (a continuous-state machine) to compute a transcendental number such as $\pi$ up to a certain digit. Turing gave this as a proof of concept that such a difference in nature disappears with the power of universal digital computing, given sufficient storage.

In summary, there is enough evidence to suggest that Turing varied the design of his imitation tests in response to the challenges posed by Hartree, Polanyi, and Jefferson. 
Turing's test was a response to critics. But was it also intended as a positive proposition?

In the 1990s, Turing's former student, close friend, and literary executor, Robin Gandy, wrote that Turing's 1950 paper ``was intended not so much as a penetrating contribution to philosophy but as propaganda.'' Gandy added: ``Turing thought the time had come for philosophers and mathematicians and scientists to take seriously the fact that computers were not merely calculating engines but were capable of behaviour which must be accounted as intelligent; he sought to persuade people that this was so. He wrote this paper unlike his mathematical papers quickly and with enjoyment. I can remember him reading aloud to me some of the passages always with a smile, sometimes with a giggle'' \cite[p.~125]{gandy1996}. We can now explore the effect of Turing's propaganda for machine intelligence on the other side of the North Atlantic.

\vspace{-10pt}
\section{TURING'S TEST AND EARLY AI}\label{sec:ai}
\noindent 
Claude Shannon visited Turing in Manchester in October 1950,%
\footnote{Claude E. Shannon, an oral history conducted in 1982 by Robert Price. IEEE History Center, Piscataway, NJ, USA.}
and may have returned to the United States with an offprint of Turing's 1950 paper,%
\footnote{The first reprint of Turing's `Computing Machinery and Intelligence' in the US appears to be that of James R. Newman (Ed.), \emph{The World of Mathematics}, vol. 4 (New York: Simon and Schuster), first published January 1, 1956.}
which he would cite in his ``Computers and Automata'' \cite{shannon1953} published in \emph{Proc. IRE} in October 1953. ``Rereading Samuel Butler's \emph{Erewhon} today,'' Shannon wrote, ``one finds `The Book of the Machines' disturbingly prophetic'' (p.~1235). Butler's novel \cite{butler1872} envisioned a revolution of the machines against a satire of the Victorians representing the human species. It was invoked as a dystopia in June 1949, first by a Catholic priest in a letter to \emph{The$\;$Times} published on the 14\textsuperscript{th} (p.~5), and then in an editorial, ``Mind, Machine, and Man,'' of the \emph{British$\;$Medical$\;$Journal} introducing Jefferson's article on the 25\textsuperscript{th} (pp.~1129-1130). Both urged scientists to disassociate themselves from Turing's research program. Turing responded with irony, not without a point. In his writings of 1950 and 1951 he referred to Butler's work for an image of his vision of the future of machines in nature and society \cite{goncalves2023irony}. 

By May 1953, Shannon was working with John McCarthy on their collection \emph{Automata Studies} \cite{mccarthy1956}, which revolved around ``the theory of Turing machines'' (p.~vii), and to which they invited Turing to contribute.%
\footnote{Shannon and McCarthy to Turing, May 18, 1953. Alan Turing Papers (Additional), University of Manchester Library, GB133 TUR/Add/123.} 
Turing declined the invitation, saying that he had been working for the last two years on ``the mathematics of morphogenesis,'' although he expected ``to get back to cybernetics very shortly.''%
\footnote{Turing to Shannon, June 3, 1953 (\emph{ibid.}).}
One year and four days later, Turing was dead, and early AI would not note his biological turn. 
Commenting on ``the Turing definition of thinking'' (p.~vi), McCarthy and Shannon found it ``interesting'' because it ``has the advantages of being operational or, in the psychologists' term, behavioristic $\hdots$ No metaphysical notions of consciousness, ego and the like are involved.'' They also thought that this very  strength could be a weakness, because it has ``the disadvantage'' of being susceptible to a memorizing machine playing the imitation game by looking up ``a suitable dictionary.''

McCarthy and Shannon referred interchangeably to ``definition'' and to a word that Turing actually used, ``criterion:''  
``While certainly no machines at the present time can even make a start at satisfying this rather strong \emph{criterion}, Turing has speculated that within a few decades it will be possible to program general purpose computers in such a way as to satisfy this test'' \cite[p.~v, emphasis added]{mccarthy1956}. 

In 1955, before the publication of \emph{Automata Studies}, McCarthy and Shannon, together with Marvin Minsky and Nathaniel Rochester, co-authored their well-known ``Proposal'' for AI research \cite{mccarthy1955}. Unlike Turing himself, they seem to have thought of machine intelligence in complete continuity with Turing machines, as their opening paragraph suggests: ``The study is to proceed on the basis of the conjecture that every aspect of learning or any other feature of intelligence can in principle be so precisely described that a machine can be made to simulate it.'' 
Turing's postwar view, which we have seen in connection with his concepts of machine and imitation above, seems instead to be that machines would learn their behavior primarily from experience, growing in intelligence like a human child, not always by being given precise instructions on how to do things.  
In any case, the proponents of the Dartmouth seminar did follow Turing closely in writing: ``For the present purpose the artificial intelligence problem is taken to be that of making a machine behave in ways that would be called intelligent if a human were so behaving'' (p.~7). This definition --- compare it with ``the Turing definition of thinking'' --- would stay. Intelligent machines would be machines designed to match or exceed human performance in a range of cognitive tasks. The implication, emphasized by both Turing and Norbert Wiener 
but not by the founders of AI, was that humans could no longer be needed for most white-collar jobs in the labor market.

In the early 1960s, E. Feigenbaum and J. Feldman noted in \emph{Computers and Thought} \cite{feigenbaum1963} that Turing's 1950 paper ``appeared five years before concrete developments in intelligent behavior by machine began to occur;'' and ``yet,'' they continued, ``it remains today one of the most cogent and thorough discussions in the literature on the general question ``Can a machine think?'' (pp.~9-10). They observed Turing's ``behavioristic posture relative to the question,'' which ``is to be decided by an \emph{unprejudiced comparison} of the alleged `thinking behavior' of the machine with normal `thinking behavior' in human beings'' (emphasis added). They concluded: ``He proposes an experiment  --- commonly called `Turing's test' --- in which the unprejudiced comparison could be made $\hdots$ Though the test has flaws, it is the best that has been proposed to date.'' 

Minsky, in the preface to his 1967 collection \cite{minsky1968}, reiterates the definition of AI as ``the science of making machines do things that would require intelligence if done by men'' (p.~v). 
Around the same time, Minsky collaborated with Stanley Kubrick and Arthur Clarke on their 1968 screenplay, also written as a novel, \emph{2001: A Space Odyssey} \cite{clarke1968}, which featured a futuristic computer named HAL: 

\begin{quotation}
\noindent
Whether HAL could actually think was a question which had been settled by the British mathematician Alan Turing back in the 1940s. Turing had pointed out that, if one could carry out a prolonged conversation with a machine --- whether by typewriter or microphone was immaterial --- without being able to distinguish between its replies and those that a man might give, then the machine was thinking, by any sensible definition of the word. HAL could pass the Turing test with ease.
\end{quotation}

\noindent
The ``Turing definition of thinking" was to become legendary.

\sethlcolor{yellow}
McCarthy and Shannon's memorizing machine objection was studied in depth by Stuart Shieber, who elaborated on its assumptions and concluded that it is invalid \cite{shieber2014}. 
But McCarthy's concept of memorizing may have been more elastic, as his later comment on Deep Blue's defeat of Gary Kasparov suggests \cite{mccarthy1997}. 
He expressed disappointment that it was mostly an achievement of computational power rather than thinking, and gave a clear argument why he thought so. Essentially, computer chess advanced by replacing heuristic techniques, which relied on the expertise of human players to prune the search space of possible moves, with brute force computing. ``[I]t is a measure of our limited understanding of the principles of artificial intelligence,'' McCarthy wrote, ``that this level of play requires many millions of times as much computing as a human chess player does.'' It may be, but that the problem was ``largely a quantitative matter'' was hinted at by Turing in his letter of c. June 1951 (Box~3). 

Ten years after Deep Blue vs. Kasparov, McCarthy referred to Turing's 1947 lecture (\emph{op. cit.}) 
as ``the first scientific discussion of human level machine intelligence,'' and to Turing's 1950 paper as ``amplifying'' that discussion into a ``goal'' \cite[p.~1174]{mccarthy2007}.

In 1992, Minsky co-authored a work of fiction, \emph{The Turing Option} (Warner, New York), in which Turing's test is featured in the preface. In 1995, Minsky took a stand against Loebner's Weizenbaum experiments, pleading to ``revoke his stupid prize, save himself some money, and spare us the horror of this obnoxious and unproductive annual publicity campaign.''%
\footnote{`Annual Minsky Loebner Prize Revocation Prize 1995 Announcement,' 2 March 1995. Available at: \url{https://groups.google.com/g/comp.ai/c/dZtU8vDD_bk/m/QYaYB18qAToJ}. Accessed 25 Nov 2023.}
In 2013, when asked about the Turing test in a taped interview, Minsky said: ``The Turing test is a joke, sort of, about saying `A machine would be intelligent if it does things that an observer would say must be being done by a human' $\hdots$ it was suggested by Alan Turing as one way to evaluate a machine but he had never intended it as being the way to decide whether a machine was really intelligent.''%
\footnote{`Marvin Minsky on AI: the Turing test is a joke!', from 23' 35'' to 24'45''. Available at \url{https://www.singularityweblog.com/marvin-minsky/}. Accessed Jun. 10, 2024.}
This materially connects McCarthy et al.'s definition of ``the AI problem'' with Turing's test, if material evidence were still needed.

Overall, it seems that all of these AI pioneers understood and were inspired by Turing's test at the level of conceptual foundations. Even if some of them also used the term ``experiment,'' none of them took it literally as a practical experiment, which would indeed imply an astonishing lack of imagination on their part. 
The Turing test helped move the burgeoning field of AI away from unproductive debates about the meaning of words, allowing Minsky, for example, to write in 1967 \cite{minsky1967}: ``Turing discusses some of these issues in his brilliant article, `Comput­ing Machines and Intelligence' [\emph{sic}], and I will not recapitulate his arguments $\hdots$ They amount, in my view, to a satisfactory refutation of many such objections'' (p.~107).

The value of Turing's test is that it has long been and still is a unifying ``definition,'' a ``criterion,'' a ``goal'' for, in the words of McCarthy et al., the science and engineering of ``making a machine behave in ways that would be called intelligent if a human were so behaving.'' 
For better or worse, every time AI succeeds in automating a new task that was once reserved for humans because it requires intelligence, ``the Turing definition of thinking'' conquers new territory, and the significance of Turing's early message to his contemporaries becomes clearer.

\section{CONCLUSION}\label{sec:conclusion}
This paper presented a new perspective on Turing's test.  
New light has been shed on Turing's concept of imitation, suggesting that it does not give a license for deception in AI. Rather, imitation was for Turing a mathematical concept, largely in continuity with his 1936 paper, although he later generalized how it could be achieved. It was also suggested that Turing's presentation of the various versions of his test fits what Mach called ``the basic method of thought experiments'' in the history of science. The historical conditions of Turing's proposal were reconstructed, showing that the basic idea of a gender test had been raised originally by Jefferson, and Turing's imitation game comes out of that context. Conversational performance was also a concession to his opponents, and overall Turing's test was a response to critics. But Turing also took the opportunity to promote his positive views. The known primary and secondary sources indicate that he became actively engaged in machine intelligence propaganda, and it was also in this spirit that he proposed his test, hoping to influence contemporaries and future generations of scientists. The question of the value of Turing's test and its relation to early AI was revisited, arguing that ``the Turing definition of thinking'' provided McCarthy, Minsky, and others with a definition of the AI problem at the level of conceptual foundations that arguably still drives AI research today. 

But whatever its utility, we can now appreciate that there is more to the imitation game. With its structural elements neatly designed as lighthearted concessions to opponents, and at the same time able to demonstrate the power of digital computing as early as 1950,  Turing's test has secured its place as one of the most beautiful thought experiments in the history of science.

\section{ACKNOWLEDGMENTS}
The author thanks Andrew Hodges, Jim Miles, and H. V. Jagadish for their valuable comments on an earlier version of this article; Mark Priestley for the gift of the Turing letters to Worsley; Fabio Cozman and Murray Shanahan for their support; The author is solely responsible for the accuracy of this work. 
The author thanks the Center for Artificial Intelligence (C4AI-USP) and the support from the São Paulo Research Foundation (FAPESP grants nos. 2019/07665-4, 2019/21489-4, and 2022/16793-9) and from the IBM Corporation. This article is a result of the project ``The Future of Artificial Intelligence: The Logical Structure of Alan Turing's Argument'').

\def\refname{BIBLIOGRAPHY}

\printbibliography

\noindent
\textbf{Bernardo Gonçalves} is currently a researcher at the Center for Artificial Intelligence (C4AI), University of São Paulo, Brazil, and a Visiting Fellow at King's College, University of Cambridge, UK. He works on Alan Turing, AI and computer science. He received Ph.D. degrees in Philosophy from the University of São Paulo and in Computational Modeling from the National Laboratory for Scientific Computing, Brazil.

\end{document}